\documentclass{lancetdh} % wraps elsarticle + your Lancet-style setup
\journal{Lancet Digital Health}
\usepackage[left, pagewise]{lineno}

\vfuzz \hfuzz
\begin{document}
%\doublespacing
%\linenumbers
%\modulolinenumbers[1]
%\resetlinenumber[1]

\begin{frontmatter}
\title{\texorpdfstring{\sffamily\bfseries\raggedright Associations Between Echocardiographic Traits and AI--ECG Predictions of Heart Failure}}
\author{Elias Stenhede\textsuperscript{1,2}}
\ead{elias.stenhede@ahus.no}

\author{Eivind Bjørkan Orstad\textsuperscript{3,2}}\author{Torbjørn Omland\textsuperscript{4,3}}\author{Henrik Schirmer\textsuperscript{3,4}}\author{Arian Ranjbar\textsuperscript{1}}

\address{\textsuperscript{1}Medical Technology \& E-Health, Akershus University Hospital, 1478 Lørenskog, Norway}
\address{\textsuperscript{2}Faculty of Medicine, University of Oslo, 0372 Oslo, Norway}
\address{\textsuperscript{3}Department of Cardiology, Akershus University Hospital, 1478 Lørenskog, Norway}
\address{\textsuperscript{4}Institute of Clinical Medicine, Campus Ahus, University of Oslo, 0317 Oslo, Norway}

\begin{abstract}
\purpleparagraph{Background}
Artificial intelligence-enabled electrocardiography (AI--ECG) can detect heart failure (HF), including disease not captured by left ventricular ejection fraction (LVEF), but the cardiac phenotypes underlying model predictions remain unclear. We therefore investigated whether AI--ECG-predicted HF risk aligns with established echocardiographic measures of myocardial dysfunction, remodelling, and filling pressures.

\purpleparagraph{Methods}
We retrospectively analysed ECG and echocardiography data from \num{8147} patients who underwent both examinations within three days at Akershus University Hospital between 1 January 2023 and 1 June 2025. A previously validated AI--ECG model for HF detection was applied to all ECGs. Spearman's rank correlation $\boldsymbol{\rho}$ quantified associations between echocardiographic parameters and AI--ECG risk. Subgroup analyses were performed by sex and left ventricular ejection fraction (LVEF). External validation included \num{36286} ECG-echocardiography pairs from Columbia University Irving Medical Center.

\purpleparagraph{Results}
Global longitudinal strain (GLS) showed the strongest correlation ($\boldsymbol{\rho=}$\,\num{0.57}), followed by mitral annular plane systolic excursion (MAPSE) ($\boldsymbol{\rho=}$\,\num{-0.49}) and LVEF ($\boldsymbol{\rho=}$\,\num{-0.45}). In patients with LVEF>50\%, correlations remained substantial for GLS, MAPSE, and diastolic-related parameters. Volumetric left ventricular indices correlated less strongly in women, whereas diastolic indices showed stronger correlations in women than in men.

\purpleparagraph{Conclusion}
Physiological validation showed that AI--ECG HF risk predictions align primarily with measures of systolic function, particularly global longitudinal strain, while also capturing diastolic-related abnormalities in patients with preserved LVEF. This approach may improve clinical interpretability and identify opportunities for model refinement.
%\purpleparagraph{Conclusion}
%AI--ECG-predicted HF risk aligned most strongly with measures of longitudinal systolic dysfunction, particularly global longitudinal strain, and remained associated with diastolic and remodelling-related indices in patients with preserved LVEF. These findings suggest that AI--ECG captures clinically relevant myocardial dysfunction beyond conventional volumetric measures.

\end{abstract}

\end{frontmatter}
\newpage

\twocolumn[
%\begin{researchinsummary}
%  \paragraph{What is known?}
%  AI-based interpretation of ECGs can aid in detecting heart failure, but the cardiac features driving these predictions remain unknown. 
%
%  \paragraph{What does this study add?}
%  We show that AI--ECG-based heart failure risk primarily correlates with left ventricular systolic function, particularly global longitudinal strain. In patients with preserved left ventricular ejection fraction, ECG risk further correlates with remodelling and diastolic parameters, particularly in women.
%
%  \paragraph{Clinical implications}
%  These findings can inform clinical personnel, ensuring an adequate level of trust in AI-based ECG interpretations. The findings also raise the question of whether emergent sex differences are due to underlying physiological differences or unwanted biases.
%\end{researchinsummary}
]

\section*{Introduction}
    Cardiovascular disease is the leading cause of death globally~\cite{stark_global_2025}, and heart failure (HF) is a common end-stage of most cardiovascular disease. HF is severely underdiagnosed due to the relatively high diagnostic complexity and low specificity of symptoms such as fatigue and dyspnea. Current diagnostic algorithms include risk factors, signs, and levels of N-terminal pro-B-type natriuretic peptide (NT-proBNP) to qualify for echocardiographic examination to rule out HF or determine its phenotype. In the last few years, AI-interpreted ECGs (AI--ECG) have emerged as a promising candidate to improve diagnostics~\cite{myhre_digital_2024}. AI--ECG models can detect a range of cardiovascular diseases, including HF~\cite{sau_artificial_2024} and its associated phenotypes~\cite{poterucha_detecting_2025, lee_artificial_2024, attia_screening_2019}, but are often perceived as black-box models by clinicians.
    
    Several post-hoc explainability techniques have been proposed and used on the ECG, such as Gradient-weighted Class Activation Mapping, Integrated Gradients, and Layer-wise Relevance Propagation~\cite{sau_artificial_2024, wagner_explaining_2024, strodthoff_deep_2021, buscher_deep_2025, czerwinski_interpretable_2025}. While these methods may highlight salient features in the ECG, their reliability is limited~\cite{arends_signal_2026}, and they provide little mechanistic insight into cardiac function. 
    
    To address this limitation, we investigate the use of established echocardiographic parameters as predictors for AI--ECG model outputs for HF detection. This approach enables the identification of HF phenotypes for which the model performs particularly well or poorly, bridging the gap between deep learning predictions and measurable cardiac physiology.

    \subsection*{Echocardiography}
    Echocardiography is a cornerstone in the evaluation of patients with suspected HF, given its widespread availability and ability to provide a real-time assessment of cardiac structure and function~\cite{lang_recommendations_2015, nagueh_recommendations_2025}. Several echocardiographic parameters provide insight into systolic performance, intracardiac pressures, and volumes, as well as overall cardiac mechanics. The indices included in the present study are summarised in \Cref{tab:echo_params_role} and described below.
    
    \begin{table*}[h]
    \centering
    \begin{threeparttable}
    {\setlength{\tabcolsep}{20pt}
    \caption{Echocardiographic parameters considered in this study and their diagnostic roles.}
    \label{tab:echo_params_role}
    \begin{tabular}{l l l}
        \toprule
        Abbreviation & Full name & Diagnostic role \\
        \midrule
        GLS & Global longitudinal strain & Left ventricular systolic function \\
        MAPSE & Mitral annular plane systolic excursion & Left ventricular systolic function \\
        LVEF & Left ventricular ejection fraction & Left ventricular systolic function \\
        TAPSE & Tricuspid annular plane systolic excursion & Right ventricular systolic function \\
        CI & Cardiac index & Global pump performance \\
        E/e' & E/e' ratio & Estimate of LV filling pressures \\
        e' & Average e' velocity & Diastolic function \\
        E/A & Mitral E/A ratio & Diastolic function \\
        A & Mitral A-wave velocity & Diastolic function \\
        LAVi & Left atrial volume index & Chronic elevation of LV filling pressures / diastolic burden \\
        IVS & Interventricular septal thickness & Left ventricular hypertrophy \\
        PWT & Posterior wall thickness & Left ventricular hypertrophy \\
        LVMi & Left ventricular mass index & Hypertrophy and remodeling assessment \\
        LVEDVi & LV end-diastolic volume index & LV dilation / chamber size assessment \\
        PASP & Pulmonary artery systolic pressure & Pulmonary hypertension \\
        \bottomrule
    \end{tabular}
    \begin{tablenotes}\footnotesize
    \item Diagnostic roles follow established echocardiographic interpretation guidelines~\cite{lang_recommendations_2015}.
    \end{tablenotes}
    }
    \end{threeparttable}
\end{table*}

    Left ventricular ejection fraction (LVEF) is calculated as stroke volume divided by end diastolic volume and remains, despite its well-known limitations, the \textit{lingua franca} for phenotyping HF across populations~\cite{lyon_2022_2022, mcdonagh_2021_2021}. HF is conventionally classified into three groups based on LVEF: reduced (<40\%), mildly reduced (41-49\%), and preserved ($\geq$50\%). This LVEF-based classification accurately identifies advanced systolic HF, but its ability to distinguish cases declines when LVEF is mildly reduced or within the normal range~\cite{smiseth_phenotyping_2023}. 
    
    Global longitudinal strain (GLS) has, over the past two decades, emerged as a more sensitive and reproducible parameter, capable of detecting myocardial dysfunction even when LVEF is within the normal range, as seen in conditions such as diastolic HF, hypertrophic cardiomyopathy, and amyloidosis~\cite{smiseth_myocardial_2016, nagueh_recommendations_2025}. Complementary indices of longitudinal performance are the mitral and tricuspid annular plane systolic excursion (MAPSE/TAPSE), which are obtainable in almost all patients and reflect basal systolic shortening in the left and right ventricles, respectively. MAPSE correlates closely with both LVEF and GLS, but its added value becomes limited in the context of more global and advanced indices of LV systolic function~\cite{stoylen_relation_2018, hu_clinical_2013}. For the right ventricle, TAPSE remains the main index for systolic function, with established prognostic value~\cite{lang_recommendations_2015, mukherjee_guidelines_2025}.

    To complement volumetric and deformation indices, Doppler-derived cardiac output (CO, stroke volume multiplied by heart rate) and cardiac index (CI, CO divided by body surface area) provide measures of ventricular performance and can help identify reduced systolic function even with a normal LVEF~\cite{mcdonagh_2021_2021}. 
    
    Diastolic function is assessed using a complementary set of echocardiographic indices reflecting left ventricular relaxation, compliance, and filling pressures. These include transmitral inflow parameters (E and A velocities and the E/A ratio), early diastolic mitral annular velocities (e'), and the  E/e' ratio, which together characterise ventricular relaxation and estimate left ventricular filling pressure. Pulmonary artery systolic pressure (PASP) reflects right ventricular afterload and is closely associated with left atrial pressure in the absence of pulmonary disease and right-sided valvular stenosis. Indexed left atrial volume (LAVi) serves as an integrated marker of chronically elevated filling pressures and long-standing diastolic dysfunction. Collectively, these indices form the echocardiographic basis for evaluating left ventricular filling pressures and grading diastolic dysfunction, which are pivotal to the diagnosis and phenotyping of heart failure with preserved ejection fraction (HFpEF)~\cite{smiseth_phenotyping_2023, nagueh_recommendations_2025}. 

    Finally, structural remodelling of the left ventricle (LV) reflects progressive myocardial adaptation to genetic or acquired influences and the cumulative effect of altered loading conditions over time. Septal and posterior wall thickness (IVS/PWT), LV mass, and indexed end-diastolic volume (LVEDi) are among the key echocardiographic indices used to characterise these processes~\cite{lang_recommendations_2015, heymans_dilated_2023}.

\section*{Methods}
    This is a retrospective observational cohort study, reported in accordance with the STROBE guidelines, and approved by the Norwegian Directorate of Health (Helsedirektoratet; case 24/45684-5; dated January 24, 2025). The study includes patients admitted to Akershus University Hospital (Ahus) between January 1, 2023, and June 1, 2025. Patients were eligible for inclusion if they had at least one standard 12-lead ECG and one transthoracic echocardiographic (TTE) examination performed within three days during the study period. For each patient, the ECG closest in time to the echocardiogram was selected for analysis. Ahus serves as a tertiary care centre within a healthcare network covering approximately 600,000 inhabitants, representing both urban and suburban populations. 
    
    Digital ECGs and echocardiograms were retrieved from the hospital's picture archiving and communication system, where they are stored using the Digital Imaging and Communications in Medicine (DICOM) standard. A previously validated AI--ECG model for HF detection was used to generate a continuous HF risk score for each ECG~\cite{stenhede_heart_2026}. The model achieved an area under the receiver operating characteristic curve of between 0.86 and 0.96, depending on target definition. The model output score ranges from 0 to 1, with higher values indicating stronger model-estimated signs of HF. The model was trained using a combination of diagnostic labels derived from ICD-10 codes and NT-proBNP, without using echocardiography. ECGs were preprocessed to the appropriate format for inference, including resampling to \SI{500}{\hertz} and per-lead z-normalisation.
    
    Because echocardiographic data were pragmatically collected as part of routine clinical care, only parameters annotated in standard reports were available for analysis. The annotated parameters varied per examination, depending on image quality and clinical indication. A subset of relevant echocardiographic parameters was selected based on relevance within HF diagnostics. Volumes, masses, and cardiac output were indexed to body surface area (BSA), estimated using the Du Bois formula~\cite{du_bois_clinical_1916}, which is the standard method applied in routine echocardiographic reporting at the hospital.

    For external validation, the model was evaluated on the EchoNext dataset from Columbia University Irving Medical Center (CUIMC), comprising paired ECG-TTE examinations collected during routine clinical care between 2008 and 2022. Patients were included if an ECG and TTE were performed within one year, and only the closest pair per patient was retained, yielding \num{36286} ECG-TTE pairs from \num{36286} patients. These ECGs were resampled to \SI{500}{\hertz}, and per-lead z-normalised; the signals were clipped at the \num{0.1}st and \num{99.9}th percentiles. Five echocardiographic parameters were included: visually determined LVEF, PASP, Tricuspid Regurgitation maximum velocity (TRv), IVS, and PWT. No retraining or calibration of the AI--ECG model was performed.
    
    In both the internal and external test cohorts, pairwise associations between the AI--ECG HF probability and each echocardiographic parameter were quantified using Spearman’s rank correlation. Rank correlation was chosen as the variables were expected to be monotonically but not linearly associated. Confidence intervals were estimated using the bootstrap percentile method, and significance levels with permutation testing, each based on \num{10000} resamples. To visualise trends, the median AI--ECG HF probability was estimated across the distribution of each echocardiographic parameter using kernel smoothing, with shaded bands representing the interquartile range. Statistical analyses were performed in Python~3.12 using \texttt{numpy~2.2.6}, \texttt{pandas~2.3.0}, and \texttt{scipy~1.16.0} libraries. The full code, including code to reproduce the results on the external cohort, is shared at \href{https://github.com/Ahus-AIM/Associations-Between-Echocardiographic-Traits-and-AI-ECG-Predictions-of-Heart-Failure}{github.com/Ahus-AIM/Associations-Between-Echocardiographic-Traits-and-AI-ECG-Predictions-of-Heart-Failure}.
    
\section*{Results}
\begin{figure}
        %\lalign
        \resizebox{0.92\linewidth}{!}{\input{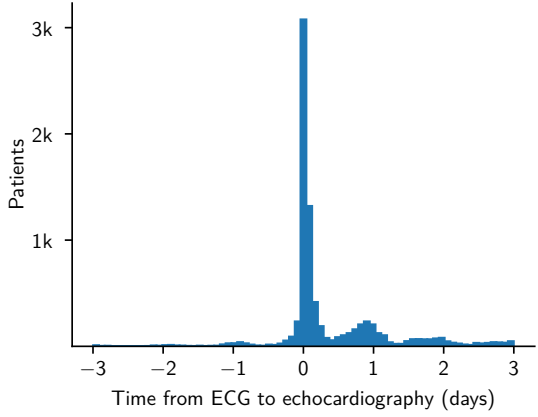}}
        \caption{Shows the relative time difference between matched ECG and echocardiographic examination for each patient in the internal Ahus test cohort}
        \label{fig:echo-ecg-time}
    \end{figure}

    In total, \num{8147} patients (42\% females) with a mean age of \num{66.5} years were included. The mean absolute time difference between ECG and echocardiogram was \num{11.2} hours, and ECG was usually recorded first. The full distribution of time differences is shown in \Cref{fig:echo-ecg-time}. Additional baseline clinical information is presented in \Cref{tab:baseline_full_echo}. Within the group, \num{735} (9\%) of the patients had been previously diagnosed with HF, and \num{2293} (28\%) diagnosed at any point before data extraction from the electronic health records. Out of these, \num{2007} patients had an annotated LVEF value in connection to ECG: 587 patients (29\%) had LVEF<40\%, 542 patients (27\%) had LVEF in 41-49\%, and 878 patients (44\%) had LVEF>50\%.

    \begin{table}[h]
	\centering
	\caption{Baseline characteristics for the full Ahus echocardiography cohort: age, sex, and comorbidities recorded before the index date and across all available time.}
	\label{tab:baseline_full_echo}
	\begin{threeparttable}
    {\setlength{\tabcolsep}{17pt}
	\begin{tabular}{@{}l rr@{}}
	\toprule
	Characteristic & At index & All time \\
	\midrule
	Patients, n & \num{8147} &  \\
	Age &  &  \\
	\quad Years, mean (SD) & \num{66.5} (\num{15.6}) &  \\
	\quad Years, median [IQR] &  \num{69} [\num{57}, \num{78}] &  \\
	Sex &  &  \\
	\quad Female & \num{3431} (\num{42}\%) &  \\
	\quad Male & \num{4713} (\num{58}\%) &  \\
	Comorbidities &  &  \\
	\quad Hypertension & \num{2073} (\num{25}\%) & \num{2966} (\num{36}\%) \\
	\quad Ischaemic heart disease & \num{1845} (\num{23}\%) & \num{3149} (\num{39}\%) \\
	\quad Atrial fibrillation & \num{1320} (\num{16}\%) & \num{2476} (\num{30}\%) \\
	\quad Myocardial infarction & \num{1078} (\num{13}\%) & \num{1776} (\num{22}\%) \\
	\quad Heart failure & \num{735} (\num{9}\%) & \num{2293} (\num{28}\%) \\
	\quad COPD & \num{530} (\num{7}\%) & \num{732} (\num{9}\%) \\
	\bottomrule
	\end{tabular}
	\begin{tablenotes}[para]
	\footnotesize
	Before index includes diagnostic codes registered before the first ECG; All time includes diagnoses recorded after the index.
	\end{tablenotes}
    }
	\end{threeparttable}
\end{table}

    The relationship between the model-predicted risk of HF and echocardiographic parameters was assessed using Spearman’s rank correlation, presented in \Cref{tab:spearman_corr}. The most correlated indices, GLS ($\rho=$\ \num{0.57}, (\num{0.54}, \num{0.59})), MAPSE ($\rho=$\ \num{-0.49}, (\num{-0.51},\num{-0.47})), and LVEF ($\rho=$\ \num{-0.45}, (\num{-0.47}, \num{-0.43})) are all measures of left ventricular systolic function. They were followed by LAVi, PASP, LVMi, E/e', and e'. High LAVi and LVMi indicate LA dilation and LV hypertrophy, and PASP, E/e', and e' are measures of elevated filling pressures and diastolic HF. Finally, TAPSE was inversely correlated with model-predicted risk of HF ($\rho=$\ \num{-0.34}, [\num{-0.36}, \num{-0.32}]). The other parameters showed weaker correlations ($|\rho|\leq$\ \num{0.22}), suggesting no strong, monotonic relationship within the full patient group.
    
    \begin{table}[h]
   \centering
   \begin{threeparttable}
   \caption{Spearman rank correlation between AI-ECG HF risk prediction and echocardiographic parameters, calculated on the Ahus cohort.}
   \label{tab:spearman_corr}
   {\setlength{\tabcolsep}{12pt}
   \begin{tabular}{lcrrr}
      \toprule
      Parameter & $n$ & $\rho$ &  \multicolumn{2}{c}{95\% CI} \\
      & & & Low & High \\
      \midrule
      GLS & \num{2899} & \num{0.57} & \num{0.54} & \num{0.59} \\
      MAPSE & \num{4177} & \num{-0.49} & \num{-0.51} & \num{-0.47} \\
      LVEF & \num{6518} & \num{-0.45} & \num{-0.47} & \num{-0.43} \\
      LAVi & \num{4936} & \num{0.43} & \num{0.41} & \num{0.45} \\
      PASP & \num{5273} & \num{0.40} & \num{0.37} & \num{0.42} \\
      LVMi & \num{4863} & \num{0.37} & \num{0.35} & \num{0.40} \\
      E/e' & \num{4426} & \num{0.36} & \num{0.34} & \num{0.39} \\
      e' & \num{7203} & \num{-0.35} & \num{-0.37} & \num{-0.33} \\
      TAPSE & \num{7248} & \num{-0.34} & \num{-0.36} & \num{-0.32} \\
      IVS & \num{2530} & \num{0.22} & \num{0.19} & \num{0.26} \\
      PWT & \num{2500} & \num{0.22} & \num{0.18} & \num{0.26} \\
      LVEDVi & \num{5179} & \num{0.14} & \num{0.12} & \num{0.17} \\
      CI & \num{5175} & \num{-0.06} & \num{-0.09} & \num{-0.03} \\
      A & \num{6853} & \num{0.06} & \num{0.03} & \num{0.08} \\
      E/A & \num{6846} & \num{-0.00} & \num{-0.03} & \num{0.02} \\
      \bottomrule
   \end{tabular}
   }
   \begin{tablenotes}\footnotesize
   \item $n$ is the number of non-missing (parameter, ensemble) pairs used for each correlation. Higher absolute $\rho$ indicates a stronger monotonic association. Values are 95\% bootstrap confidence intervals (\num{10000} resamples).
   \end{tablenotes}
   \end{threeparttable}
\end{table}

    \begin{table}[h]
   \centering
   \begin{threeparttable}
   \caption{Spearman rank correlation between parameters and AI-ECG risk of HF by LVEF subgroup and sex, calculated on the Ahus cohort.}
   \label{tab:spearman_corr_lvef_sex}
   {\setlength{\tabcolsep}{4pt}
   \begin{tabular}{lrrrrrr}
      \toprule
      & \multicolumn{3}{c}{LVEF subgroup} & \multicolumn{3}{c}{Sex} \\
      \cmidrule(lr){2-4}
      \cmidrule(lr){5-7}
       & $\leq$ 40\% & 41--49\% & $\geq$ 50\% & Male & Female & p \\
      \midrule
      GLS & \cellcolor{green!31} \num{0.44} & \cellcolor{green!16} \num{0.34} & \cellcolor{green!21} \num{0.37} & \cellcolor{green!50} \num{0.57} & \cellcolor{green!45} \num{0.54} & \num{0.252} \\
      MAPSE & \cellcolor{green!3} \num{-0.25} & \cellcolor{green!10} \num{-0.30} & \cellcolor{green!17} \num{-0.35} & \cellcolor{green!38} \num{-0.48} & \cellcolor{green!43} \num{-0.52} & \num{0.185} \\
      LVEF & \cellcolor{green!5} \num{-0.26} & \cellcolor{green!0} \num{-0.20} & \cellcolor{green!0} \num{-0.16} & \cellcolor{green!38} \num{-0.49} & \cellcolor{green!24} \num{-0.39} & <\num{0.001} \\
      LAVi & \cellcolor{green!14} \num{0.32} & \cellcolor{green!16} \num{0.33} & \cellcolor{green!20} \num{0.36} & \cellcolor{green!27} \num{0.41} & \cellcolor{green!31} \num{0.44} & \num{0.265} \\
      PASP & \cellcolor{green!5} \num{0.26} & \cellcolor{green!10} \num{0.29} & \cellcolor{green!20} \num{0.37} & \cellcolor{green!22} \num{0.38} & \cellcolor{green!30} \num{0.43} & \num{0.027} \\
      LVMi & \cellcolor{green!0} \num{0.21} & \cellcolor{green!0} \num{0.18} & \cellcolor{green!8} \num{0.28} & \cellcolor{green!18} \num{0.35} & \cellcolor{green!25} \num{0.40} & \num{0.070} \\
      E/e' & \cellcolor{green!3} \num{0.25} & \cellcolor{green!2} \num{0.24} & \cellcolor{green!19} \num{0.36} & \cellcolor{green!14} \num{0.32} & \cellcolor{green!32} \num{0.45} & <\num{0.001} \\
      e' & \cellcolor{green!0} \num{-0.20} & \cellcolor{green!0} \num{-0.18} & \cellcolor{green!12} \num{-0.31} & \cellcolor{green!19} \num{-0.36} & \cellcolor{green!19} \num{-0.36} & \num{0.875} \\
      TAPSE & \cellcolor{green!9} \num{-0.29} & \cellcolor{green!0} \num{-0.21} & \cellcolor{green!0} \num{-0.22} & \cellcolor{green!15} \num{-0.33} & \cellcolor{green!19} \num{-0.36} & \num{0.175} \\
      IVS & \cellcolor{green!0} \num{-0.05} & \cellcolor{green!0} \num{0.12} & \cellcolor{green!4} \num{0.26} & \cellcolor{green!0} \num{0.11} & \cellcolor{green!17} \num{0.34} & <\num{0.001} \\
      PWT & \cellcolor{green!0} \num{0.05} & \cellcolor{green!0} \num{0.09} & \cellcolor{green!5} \num{0.26} & \cellcolor{green!0} \num{0.11} & \cellcolor{green!15} \num{0.33} & <\num{0.001} \\
      LVEDVi & \cellcolor{green!0} \num{0.22} & \cellcolor{green!0} \num{0.04} & \cellcolor{green!0} \num{-0.02} & \cellcolor{green!0} \num{0.22} & \cellcolor{green!0} \num{-0.01} & <\num{0.001} \\
      CI & \cellcolor{green!0} \num{0.06} & \cellcolor{green!0} \num{0.01} & \cellcolor{green!0} \num{-0.02} & \cellcolor{green!0} \num{-0.03} & \cellcolor{green!0} \num{-0.12} & \num{0.001} \\
      A & \cellcolor{green!0} \num{-0.11} & \cellcolor{green!0} \num{-0.02} & \cellcolor{green!0} \num{0.13} & \cellcolor{green!0} \num{0.03} & \cellcolor{green!0} \num{0.12} & <\num{0.001} \\
      E/A & \cellcolor{green!0} \num{0.14} & \cellcolor{green!0} \num{0.15} & \cellcolor{green!0} \num{-0.06} & \cellcolor{green!0} \num{0.02} & \cellcolor{green!0} \num{-0.03} & \num{0.075} \\
      \bottomrule
   \end{tabular}
    }
   \begin{tablenotes}\footnotesize
   \item Higher absolute $\rho$ indicates stronger monotonic association. The $p$ column reports two-sided permutation test $p$-values for the difference in correlation between males and females, based on \num{10000} random reassignments of sex labels. Values are capped at $<\num{0.001}$.
   \end{tablenotes}
   \end{threeparttable}
\end{table}

    Subgroup analysis based on LVEF and sex was performed and presented in \Cref{tab:spearman_corr_lvef_sex}. After stratification by LVEF, GLS remained among the most strongly correlated indices. LAVi and PASP also correlated relatively well across the LVEF stratum. Diastolic-related parameters exhibited higher correlations in patients with EF >50\%. There were significant sex differences in LVEF and LVEDVi, with the model being more strongly correlated in men. Conversely, the model was significantly more correlated with E/e', IVS, PWT, CI, and A-wave velocity in women than in men.
    
\begin{figure*}[h]
            \flushright
            \resizebox{0.98\linewidth}{!}{\input{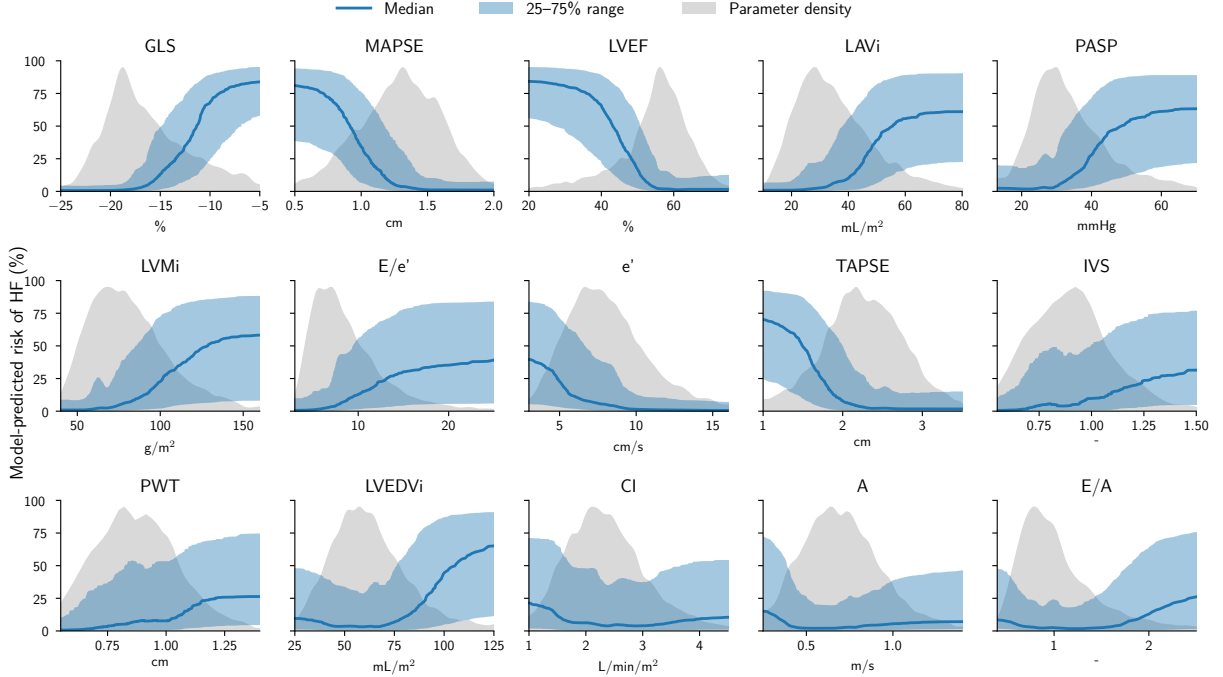}}
            \caption{Shows the median and IQR of model-predicted risk as a function of varying echocardiographic parameters in the internal evaluation dataset, sorted by Spearman's rank correlation.}
            \label{fig:rankx}
    \end{figure*}

    We plotted the kernel-estimated median of the model-predicted risk as a function of each parameter, along with the interquartile range of the data, in \Cref{fig:rankx}. Indeed, we find that the least correlated parameters often exhibited non-monotonic relationships, where model-predicted risk increased at both extremes.
    
\begin{figure*}[htpb]
        \flushright
        \resizebox{0.98\linewidth}{!}{\input{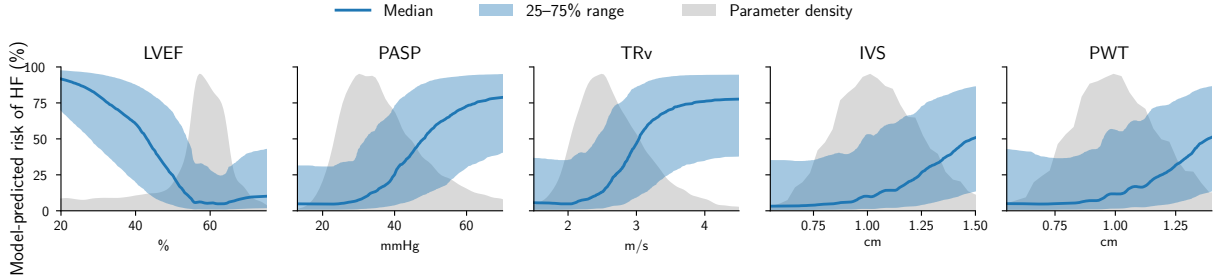}}
        \caption{Median and IQR of model-predicted risk as a function of varying echocardiographic parameters from the EchoNext dataset, collected at Columbia University Irving Medical Center.}
        \label{fig:rankx-echonext}
    \end{figure*}

    The same kernel-smoothing procedure was applied to the external EchoNext dataset, shown in \Cref{fig:rankx-echonext}, which consisted of \num{36286} ECG-TTE pairs. Compared with the internal cohort, the external dataset exhibited a higher median model-predicted risk across the full range of echocardiographic parameters. Despite these shifts, the qualitative relationships between the model output and echocardiographic traits remain similar.

\begin{figure*}[h]
        \centering
        \resizebox{\linewidth}{!}{\input{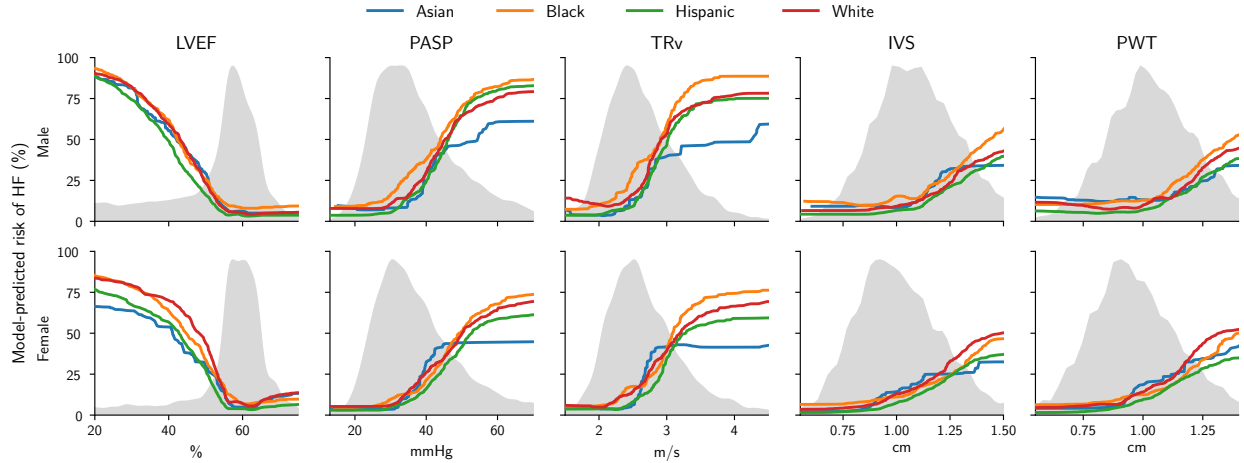}}
        \caption{Shows the median model-predicted risk stratified by echocardiographic parameters, sex, and race, from the EchoNext dataset, collected at CUIMC.}
        \label{fig:rankx-echonext-race-sex}
    \end{figure*}

\begin{figure*}[h]
        \flushright
        \resizebox{0.98\linewidth}{!}{\input{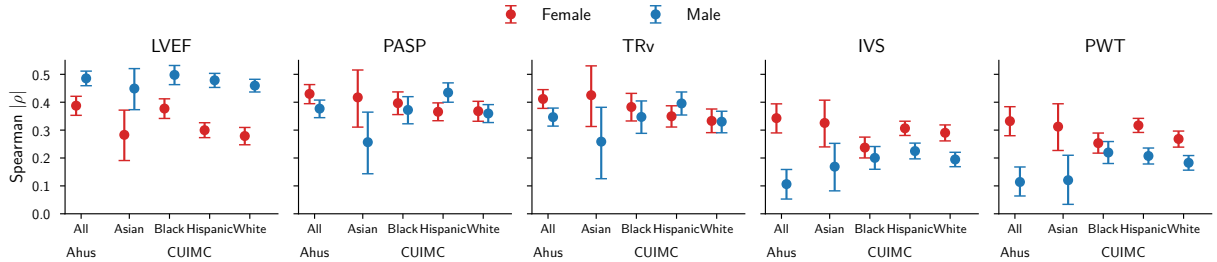}}
        \caption{Spearman's rank correlation and its 95\% symmetric confidence intervals for five echocardiographic indices, split by medical site, sex, and race.}
        \label{fig:median-race-sex}
    \end{figure*}

    In the external CUIMC cohort, race was reported for each ECG-TTE pair, enabling stratification by sex and race. The kernel-based median as a function of echocardiographic indices is shown in \Cref{fig:rankx-echonext-race-sex}, with corresponding group sizes presented in \Cref{tab:race_sex_counts}. Black men were generally assigned a higher AI--ECG risk of HF, and Asian men were estimated as relatively lower risk at abnormal echocardiographic values. It should, however, be noted that the Asian group was by far the smallest. To assess differences between the Ahus cohort and CUIMC, rank correlations of the common echocardiographic indices are presented in \Cref{fig:median-race-sex}. The LVEF correlated significantly better in men than in women across all groups. For IVS and PWT, the same directional relationship was seen in the CUIMC cohort as reported in \Cref{tab:spearman_corr_lvef_sex}, but the sex differences were generally smaller.

    \begin{table}[h]
    \centering
    \begin{threeparttable}
    \caption{Counts of in the external CUIMC cohort, stratified by race and sex.}
    \label{tab:race_sex_counts}
    {\setlength{\tabcolsep}{17pt}
    \begin{tabular}{lrrr}
        \toprule
        Race & Total & Male & Female \\
        \midrule
        Hispanic & \num{11006} & \num{4782} & \num{6224} \\
        White & \num{10635} & \num{6037} & \num{4598} \\
        Black & \num{5424} & \num{2408} & \num{3016} \\
        Unknown & \num{5159} & \num{2635} & \num{2524} \\
        Other & \num{2980} & \num{1530} & \num{1450} \\
        Asian & \num{1082} & \num{561} & \num{521} \\
        \bottomrule
    \end{tabular}
    }
    \end{threeparttable}
\end{table}

    The full correlation matrix of all parameters sorted by absolute correlation with AI--ECG-predicted HF risk is displayed in Supplementary Figure 1.
    
\section*{Discussion}
    %\subsection*{Summary of principal findings}
    The AI--ECG model's HF predictions primarily reflect LV systolic function, with GLS showing the strongest association, followed by MAPSE and LVEF in men. Among patients with LVEF>50\%, the model remained associated with GLS and MAPSE, as well as with diastolic and remodelling markers, including left atrial volume, pulmonary pressures, and the E/e' ratio. Sex-stratified analyses reveal consistently stronger correlations in women, except in LV volumetric parameters.
    
    %\subsection*{Analysis of our findings}
    The relatively weaker correlation between AI--ECG risk and LVEF than between AI--ECG risk and GLS suggests that the model may identify early systolic impairment that is not captured by volumetric measures. About half of HF patients have preserved LVEF, and systolic dysfunction in these patients is still measurable via GLS~\cite{kraigher-krainer_impaired_2014}, possibly explaining its higher correlation. The associations to diastolic and filling pressures-related parameters were most pronounced in patients with preserved LVEF, indicating that the model detects features consistent with HFpEF.

    A notable result was the difference in associations between men and women. In the group with LVEF>50\%, the model correlated more strongly with diastolic and structural parameters in women. This pattern might reflect known sex differences in HFpEF, such as women having worse exercise capacity compared to men with similar resting cardiac function~\cite{mauricio_sex_2019}. As the model appears to have learned sex-specific phenotypes, even though sex was neither a target nor a model input during training, further research is warranted to determine whether these emergent patterns are beneficial or harmful to fairness and model utility.
    
    The model output increased sharply near clinically relevant thresholds, which aligns with how systolic and diastolic dysfunction is interpreted in routine practice. Patients with GLS of more than -18\% consistently had low model-predicted risk, whereas patients with GLS of less than -16\% had increasingly higher model-predicted risk of HF. Patients with LVEF of above 55\% also had low risk-predictions, whereas patients with LVEF below 40\% had consistently high predicted risk. For PASP, values above \num{30}\ \unit{mmHg} were correlated with increasing predicted risk.
    
    Evaluation on an external EchoNext dataset collected at CUIMC showed similar relationship patterns, but with notable differences. The distribution of echocardiographic indices differed, with a much larger proportion of patients having very low LVEF. Differences likely reflect variations in patient demographics, acquisition hardware, and echocardiographic reporting; LVEF was visually estimated in the external cohort. The model generally predicted a higher risk of HF in the external cohort, but the correlation between the model's predicted risk and parameters was similar. The consistency supports the generalizability of using echocardiographic traits to interpret model behaviour at scale. Stratified by race and sex, the highest median predictions were assigned to Black males and the lowest to Asian men. The results may reflect differences in HF prevalence between groups, but this cannot be verified because diagnostic labels are not included in the CUIMC dataset.

    Current explainability methods for AI--ECG fall into two categories: inherent and post-hoc, each with limitations. Inherent methods, such as variational autoencoding-based models, offer human-interpretable latent features but cause substantial performance drops compared with less interpretable models~\cite{patlatzoglou_cost_2025, sau_artificial_2024}. Post-hoc approaches avoid this performance drop but introduce other issues: instance-level explanations can increase human agreement even with incorrect AI outputs, reinforcing confirmation bias~\cite{gomez_explainable_2025, ghassemi_false_2021, muller_how_2025}, and despite fostering trust, clinical trials have not demonstrated that explainable systems improve outcomes relative to black-box models~\cite{nicolson_human_2025}. Post-hoc methods can nevertheless be useful in specific scenarios; when models operate on paper-ECG images, saliency maps may highlight regions unrelated to the task and thereby provide partial insight. In contrast, interpreting saliency maps on digital ECGs is challenging. High saliency on any given signal segment may reflect physiological features or undesired confounding measurement noise, with no clear way to disentangle them.

    %\subsection*{Clinical implications}
    Cohort-level correlation analysis offers a practical and clinically intuitive way to interpret AI--ECG models. Identifying the echocardiographic traits most strongly associated with predictions may help clinicians understand the physiological basis of model outputs, reveal subgroups in which the model behaves differently, and guide targeted improvements in model development. Trait-based explanations may also serve an educational role, helping clinicians contextualise AI outputs within familiar cardiac physiology.
    
    %\subsection*{Limitations}
    The present study should be interpreted with its limitations in mind. First, although the sex-specific associations observed are consistent with recognised physiological differences in HFpEF, these differences are not the only possible explanations, because the AI--ECG model was trained on imperfect diagnostic labels; its outputs might reflect biases in clinical workflows present in its training data. Second, although the time difference between ECG and echocardiographic examinations was always less than three days in the Ahus cohort, changes in clinical status between tests likely led to a slight underestimation of the true correlation between AI--ECG risk score and each echocardiographic parameter. Third, echocardiographic missingness was partly driven by clinical indication; the reported correlations should be interpreted accordingly. Fourth, cohort-level explanations do not capture all the factors that drive individual predictions, and individual failure modes will not necessarily be found using the method.
    
    %\subsection*{Conclusion}
    An AI--ECG model trained on ICD-10 diagnostic labels aligns closely with established echocardiographic markers of systolic and diastolic dysfunction. This physiology-anchored, cohort-level interpretability approach provides insight into what the model detects, highlights meaningful subgroup differences, and may support more informed and equitable clinical use of AI--ECG systems. Future work should assess whether trait-based explanations facilitate clinical adoption and whether retraining targeted at physiologically weak areas can improve model utility.
    
\section*{Contributors}
    %ES and HS conceived the study. ES was responsible for data collection, anonymisation, and statistical analysis. EBO and HS provided clinical expertise and ensured the analysis's clinical relevance. ES and EBO prepared the initial draft of the manuscript. All authors contributed to the discussion, expanded and reviewed the final manuscript, and approved it for submission. All authors had full access to the data used in the study and had final responsibility for the decision to submit for publication.
    Elias Stenhede: Conceptualisation, Data curation, Formal analysis, Software, Visualisation, Writing - original draft.
    Henrik Schirmer: Conceptualisation, Clinical supervision, Methodology, Writing - review \& editing.
    Eivind Bjørkan Orstad: Clinical interpretation, Validation, Writing - original draft, Writing - review \& editing.
    Torbjørn Omland: Clinical interpretation, Writing - review \& editing.
    Arian Ranjbar: Writing - review \& editing.
    All authors approved the final manuscript and had full access to the study data.

\section*{Declaration of interests}
    Henrik Schirmer has previously received lecture fees from Amgen, Boehringer Ingelheim, Bristol-Myers Squibb, Novartis, NovoNordisk, and Sanofi-Aventis. Torbjørn Omland reports receiving personal fees from Abbott Laboratories, Roche Diagnostics, CardiNor, SpinChip and Novo Nordisk; receiving grants from Abbott Laboratories; and receiving nonfinancial support from Roche Diagnostics, Novartis, ChromaDex Inc., a Niagen Bioscience company, and CardiNor. Torbjørn Omland is supported by a grant from the Kristian Gerhard Jebsen Foundation (grant number SKGJ-MED-024). The authors declare no other competing interests.
    
\section*{Acknowledgments}
    We thank Akershus University Hospital for funding this research.

\printbibliography

\end{document}